# Duckworth-Lewis-Stern Method Comparison with Machine Learning Approach

Kumail Abbas and Sajjad Haider
Faculty of Computer Science
Institute of Business Administration (IBA)
Karachi, Pakistan
kumail.abbas@khi.iba.edu.pk, sahaider@iba.edu.pk

*Abstract*—This work presents an analysis of the Duckworth-Lewis-Stern (DLS) method for One Day International (ODI) cricket matches. The accuracy of the DLS method is compared against various supervised learning algorithms for result prediction. The result of a cricket match is predicted during the second inning. The paper also optimized DLS resource table which is used in the Duckworth-Lewis (D/L) formula to increase its predictive power. Finally, an Unpredictability Index is developed that ranks different cricket playing nations according to how unpredictable they are while playing an ODI match.

## I. INTRODUCTION

"One Day International" (ODI) matches are the most common form of limited overs cricket [5] played at the international level. ODI matches were introduced in 1971 when the first match was played between Australia and England at the Melbourne Cricket Ground (MCG). In an ODI match, each team faces a limited number of overs (typically 50) during their respective innings. The main objective of a limited-overs match is to produce a definite result and, as such, a conventional draw is not possible. The matches, however, could have undecided results if bad weather prevents a result. This is where the Duckworth-Lewis-Stern (DLS in short) method comes into the picture as it decides the winner based on the current situation of the match provided that at least 20 overs in the second inning have been bowled. DLS method was formally named as Duckworth-Lewis (D/L in short) method [7]. It was proposed in 1997 and officially adopted by the ICC in 1999. It was renamed to its current title in 2014 after Professor Steven Stern became its custodian. The professional edition of the method still uses the same D/L formula for calculation of D/L Par Score that was proposed initially but the DLS resource table has been updated according to the modern game. The D/L Professional Edition has been updated to Stern Edition in October 2014 and is now referred as Duckworth-Lewis-Stern (DLS) system. The DLS method comprises different methods which include adjusting the target or overs when the match is interrupted and then resumed afterward. In this paper, we are considering only that aspect of using D/L formula that provides D/L Par Score which decides the winner if a game is not continued after interruption. [10] It calculates the runs team batting second should have scored when the match is abandoned. If the team batting second has scored less than the D/L Par Score, the bowling team is declared the winner; otherwise the batting team wins. In this paper, we have used this Par Score attribute along with other variables to predict the outcome of an ODI match during different stages of the second inning. The classifiers presented in this paper achieve higher accuracy than the D/L formula.

Even though the ODI records are available since the very beginning, our research presents a novel approach that analyzes the utility of the DLS method in predicting the outcome of a match. It is worth mentioning that statistical research on Cricket started six to seven decades ago. In 1945, Wood [2] used Geometric distribution to model the total score in a test match. Despite the fact that the study was not based on ODI matches, it has been recognized as the pioneering research in the game of cricket. Bandulasiri [6] predicted the winner of an ODI cricket match. The work calculated the effectiveness of the DLS method for interrupted matches and used statistical methods to find winning factors of an ODI match. Bailey and Clarke [1] predicted the outcome of an ODI match while the game was in progress. They developed statistical models for prediction purposes. They proposed a formula to predict the total score of the batting team. Their work suggested how available match resources (number of overs and batsmen left) affect the final result of a match. This is similar to our work except for the fact that instead of statistical models, we are using machine learning algorithms to predict the outcome of a match.

Another similar work is that of Amal and Aparna [8], who developed a classification based tool, CricAI, to predict the outcome of an ODI match. The tool, however, only provides the winning probability of both teams even before the start of a match and does not take into account the D/L formula. They also haven't calculated the accuracy of their tool. Another work by De Silva [4] analyzes the magnitude of the victory. The paper mainly focuses on identifying the factors affecting winning and do not predict the winner of a match.

M. Bailey and S.R. Clark [1] applied multiple linear regressions to assign winning probabilities to the competing

teams in ODI matches by considering a total of 2200 ODI matches. Using a holdout sample of 100 completed matches played in the year 2005, they developed a regression model to successfully predict the winning team 71% of the time which is significantly less than 80.62% that we achieved by using Neural Networks with backpropagation.

The rest of the paper is organized as follows: Section II presents a broad overview of our methodology and how data from online sources were extracted for this analysis. Section III learns classifiers to predict the winner of a match. It also presents a Particle Swarm Optimization (PSO) based method to optimize the DLS resource table. Finally, it presents the Unpredictability Index of all national teams that demonstrates how unpredictable a team is based on their previous performances using the D/L formula. Finally, Section IV concludes the paper and provides future research directions.

## II. DATA PREPARATION

In this work, we collected data of 3,470 ODI matches played since 1971 from the CricInfo website [3]. The runs scored by the team batting second and wickets fallen after every over is also grabbed using a separate section available on the website. The section, however, is only available for ODI matches played after 7th June 2001. Thus, the total number of available records (matches) were reduced to 1,751 (3470-1719 = 1,751). We also excluded tied and rain-interrupted matches from our samples. After doing this data processing, we were left with records of 63,000 overs bowled during the second innings of ODI matches. For each extracted record, we have the following attributes available:

- Team 1 (Team batting first)
- Team 2 (Opponent team)
- Toss
- Runs scored by the team batting first
- Wickets fallen by Team 2
- Overs bowled during second innings
- Runs scored by the team batting second
- The actual result of the match

We calculated D/L Par Score for each over that we collected and introduced one more attribute in collected data "D/L Winning Prediction" (If current runs scored by team batting second is less than D/L par score than it outputs bowling team as a winner and vice versa). The process flow of grabbing and parsing the data is shown in Fig. 1.

It is worth mentioning that during the second inning of an ODI cricket match, we can calculate D/L Par Score after every delivery. This D/L Par Score declares the winner of the match if the match is abandoned at that time. While predicting the winner of a match, the D/L Par Score could also be viewed as an attribute that predicts who would win the match if the match is not abandoned at a specific over and is continued till the end. In this work, we have used this par score as an input variable while learning classifiers to predict the winner of the match. For this purpose, we have collected runs scored and wickets fallen during second innings of the match for all the matches that have this data available online. It must be noted that to avoid complexity in our analysis, we have collected the progress of a match after every over and not after every delivery. In the list of original variables mentioned above, we have added, calculated D/L Par Score and another attribute

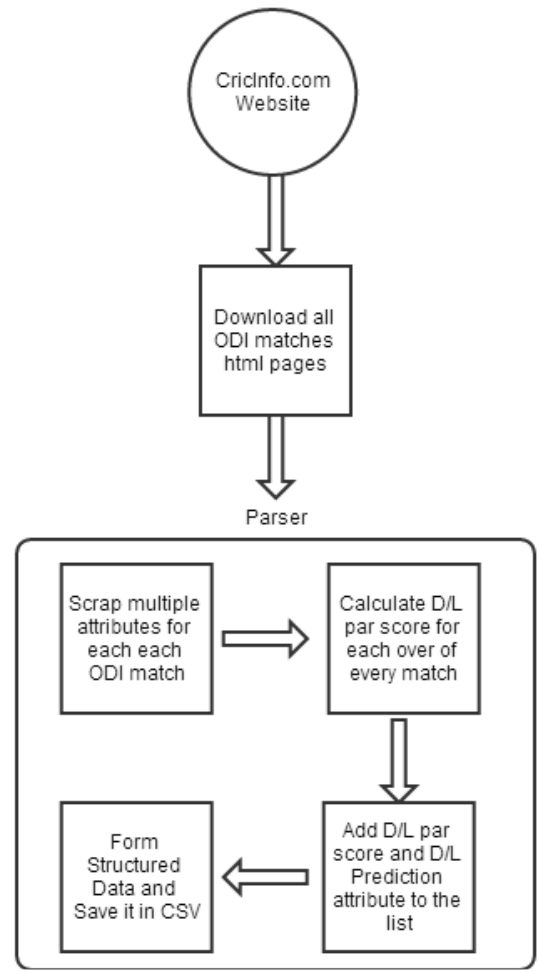

Fig. 1. Process flow of Scrapping data from CricInfo.com

"D/L Winning Prediction". The last column predicts a winner based on the runs scored by the team batting second and whether is it more or less than the par score. The complete process flow of grabbing and parsing the data is shown in Fig. 1.

## III. APPLICATION AND ANALYSIS

In this work, we have used KNIME as the main data analytics tool for analysis and model developments in all experiments except the optimization of DLS Resource Table. In the sequel, we used the grabbed/modified ODI data set to perform the following experiments:

1) Analysis of D/L Formula for better prediction (Predicting the winner of an ODI match using several attributes including the D/L Formula)
2) DLS Resource Table Optimization (Optimizing DLS Resource Table for better table values)
3) Unpredictability Index (Computing Unpredictability Index of cricket playing nations)

*A. Analysis of D/L Formula for better prediction*

The accuracy of the D/L formula was computed by simply comparing the "D/L Winning Prediction" attribute with "Actual result of the match". The accuracy of "D/L Winning Prediction" came out to be 78.12% when applied to all 63,000 overs samples. This did not seem satisfactory as this also included the starting 20 overs of the second innings for which the DLS method is not applicable. Thus, the first 20 overs were excluded from the analysis and then the accuracy reached 83.74%. In addition, the D/L formula's accuracy was also calculated at 10th, 20th, 30th and 40th overs. The results are shown in Table I (3rd column).

We also trained several classifiers to predict the winner of a match at the end of 10th, 20th, 30th and 40th overs. The list of classifiers includes Naïve Bayes, Neural Networks (NN), Bagging with Naïve Bayes and Random Forest. The comparison of the D/L formula with each classifier is also given in Table I. It is worth mentioning that almost all the classifiers achieved accuracy of around 79% when only 10 overs of the second innings were bowled which is almost equal to the D/L formula's accuracy at even 20 overs stage.

The classifiers in Table I was built in the following manner. For a particular over (say 20th over - 2nd row of Table I), records related to that over of all the completed matches were extracted. Each classifier was trained from 70% of those records while 30% were retained for testing. The values shown in Table I are of that 30% testing sample.

To further analyze and improve D/L prediction, we built our classifiers using ranges of overs. For example, in Row 2 of Table II (10-20 overs), we grabbed all the data between overs 10 and 20, built classifiers from 70% of that data and test their accuracy on the remaining 30%. The values shown in Table II reflect the accuracy of classifiers on the 30% testing data. A similar analysis was performed for larger ranges of overs, 0-50 and 20-50, and the results are shown in the second last and the last rows of Table II, respectively.

TABLE I. COMPARISON OF D/L FORMULA WITH DIFFERENT CLASSIFICATION MODELS

| Test Samples (Approx. 30%) | Overs Bowled | D/L Accuracy | Classifier Accuracy | Classification Model |
|---|---|---|---|---|
| 515 ODIs | 10th over | 72.23% | 80.97% | Bagging with Naive Bayes |
| 500 ODIs | 20th over | 79.20% | 82.80% | Neural Networks |
| 454 ODIs | 30th over | 85.90% | 87.45% | Naïve Bayes |
| 354 ODIs | 40th over | 88.42% | 88.98% | Neural Networks |

As is the case with the results of Table I, all the classifiers perform better than the DLS method. Classifiers of Table I and Table II are built using the following attributes:

- Team 1 (Team batting first) Runs
- Team 1 Wickets
- Team 2 (Team Batting Second) Runs
- Team 2 Wickets
- Overs Played by Team 2
- D/L Winning Prediction

TABLE II. COMPARISON OF D/L FORMULA WITH DIFFERENT CLASSIFICATION MODELS DURING THE RANGE OF DIFFERENT OVERS

| Test Samples (Approx. 30%) | Overs Bowled | D/L Accuracy | Classifier Accuracy | Classification Algorithm |
|---|---|---|---|---|
| 4,608 overs | 0-10 overs | 67.10% | 75.22% | Neural Networks |
| 4,984 overs | 10-20 overs | 76.04% | 79.29% | Neural Networks |
| 4,713 overs | 20-30 overs | 80.99% | 82.91% | Bagging with Naive Bayes |
| 4,065 overs | 30-40 overs | 84.74% | 85.77% | Neural Networks |
| 2,391 overs | 40-50 overs | 88.75% | 89.06% | Neural Networks |
| 19,112 overs | 0-50 overs | 78.12% | 80.62% | Neural Networks |
| 10,423 overs | 20-50 overs | 83.74% | 84.68% | Neural Networks |

*B. DLS Resource Table Optimization*

F. C. Duckworth and A. J. Lewis [7] in their paper devised a DLS Resource Table to calculate D/L Par Score at any particular instance during the second innings of a match. The formula to calculate D/L par score during the second innings when X overs are left and Y wickets have been lost is:

*Par Score = Team 1 Runs – Team 1 Runs * ResourceValue(X,Y)*

where ResourceValue(X,Y) represents the percentage of resources remaining for the second team to chase the target at a particular instant. The value is extracted from the DLS Resource Table partially shown in Fig. 2. For instance, if a team batting first scores 250 runs then D/L Par Score during second innings when 40 overs are left and four wickets are lost would be 101 (250 - 250 * 59.5%).

During our analysis, we noted that the table is less accurate for the first four wickets. We thus tried to optimize this table values for wickets 0 to 3 (during the whole 50 overs). We have applied Particle Swarm Optimization (PSO) by developing a custom software application in .NET to optimize the values of the table. The fitness function used is the number of instances correctly classified by a resource value. The other parameters of the PSO algorithms are set as:

$\mu=10, c_1=2, c_2=2.5, \text{Generations}=50$

Fig. 2. Process flow of Scrapping data from CricInfo.com

The accuracy of the optimized DLS Resource Table is compared against the accuracy of the existing DLS Resource Table (for wickets 0 to 3, during the whole 50 overs) in Table III. The result suggests that there is still room for better resource values in the DLS Resource Table. It is worth mentioning that the DLS resource table always decreases from top to bottom as the number of remaining overs decreases. In modern-day cricket, this is usually not always the case especially when wickets have fallen early. In such situations, the team batting second tries to play as many overs as they can without losing wickets even at the cost of scoring very slowly. The idea is to keep wickets in hand and accelerate towards the end overs. While optimizing the DLS resource table, we imposed this decreasing condition to comply with the original table; otherwise, we were able to achieve even higher accuracy when the condition was not complied with.

TABLE III. COMPARISON OF DLS RESOURCE TABLE WITH MODIFIED RESOURCE TABLE

| No. of Samples | Overs Bowled | D/L Resource Table (Accuracy) | Modified Resource Table (Accuracy) |
|---|---|---|---|
| 1,514 | 10th over | 73.38% | 76.15% |
| 1,468 | 20th over | 80.18% | 80.59% |
| 1,335 | 30th over | 85.01% | 85.61% |
| 1,041 | 40th over | 87.03% | 87.31% |
| 15,187 | 0-10 overs | 67.59% | 69.96% |
| 16,448 | 10-20 overs | 77.31% | 78.68% |
| 15,469 | 20-30 overs | 82.46% | 82.85% |
| 13,321 | 30-40 overs | 86.06% | 86.34% |
| 7,931 | 40-50 overs | 88.98% | 89.05% |
| 63,000 | 0-50 overs | 79.25% | 80.25% |
| 34,345 | 20-50 overs | 85.12% | 85.39% |

*C. Unpredictability Index*

For all the ODIs which were collected, we analyzed the accuracy of the D/L formula at 40th over for each team separately. The lesser the accuracy of D/L formula is, the more it fails to predict the winner and ultimately the more unpredictable the team is. We have calculated unpredictability index for test playing nations for different scenarios shown in Table IV, Table V, Table VI and Table VII. We have ignored the cases where there are less than 40 ODIs available for a specific team. Fig. 3 shows a sample for calculating unpredictability index for a chasing team (Team 2) while Fig. 4 shows a sample for calculating unpredictability index for the team defending a target (Team 1).

| Actual Result of the Match \ D/L Winning Prediction | Team 1 | Team 2 | |
|---|---|---|---|
| Team 1 | 41 | 6 | 47 |
| Team 2 (Sri Lanka) | 8 | 36 | 44 |

D/L Failing Percentage when Sri Lanka won the match while chasing = 8/44 = 18.18% (Table IV)
D/L Failing Percentage when Sri Lanka lost the match while chasing = 6/47 = 12.77% (Table V)

Fig. 3. Confusion matrix showing unpredictability index for Sri Lanka while chasing a target

| Actual Result of the Match \ D/L Winning Prediction | Team 1 | Team 2 | |
|---|---|---|---|
| Team 1 (India) | 56 | 12 | 68 |
| Team 2 | 7 | 39 | 46 |

D/L Failing Percentage when India won the match while defending = 12/68 = 17.65% (Table VI)
D/L Failing Percentage when India lost the match while defending = 7/46 = 15.22% (Table VII)

Fig. 4. Confusion matrix showing unpredictability index for India while defending a target

Table IV presents the unpredictability index of teams when they won a match while chasing a target. This shows how many times a team actually won the match while chasing when D/L formula was predicting against it at the 40th over.

This shows the hard-hitting nature of the Sri Lankan team that has won matches for them from nowhere which is the main reason why they won the Twenty20 [11] Cricket World Cup in 2014.

TABLE IV. D/L FORMULA FAILING PERCENTAGE AT 40TH OVER WHEN A TEAM WON THE MATCH WHILE CHASING A TARGET

| Team | Percentage | Rank |
|---|---|---|
| SriLanka | 18.18% | 1 |
| Pakistan | 13.79% | 2 |
| Australia | 11.11% | 3 |
| England | 11.11% | 4 |
| South Africa | 10.64% | 5 |
| India | 8.86% | 6 |

Table V presents the unpredictability index of teams when they lost a match while chasing a target. This shows how many times a team actually lost the match while chasing when the D/L formula was predicting against it at 40th over

Table V shows that India collapses more often while chasing a target than other teams.

TABLE V. D/L FORMULA FAILING PERCENTAGE AT 40TH OVER WHEN A TEAM LOST THE MATCH WHILE CHASING A TARGET

| Team | Percentage | Rank |
|---|---|---|
| India | 20.37% | 1 |
| Pakistan | 16.98% | 2 |
| South Africa | 15.00% | 3 |
| England | 13.04% | 4 |
| SriLanka | 12.77% | 5 |
| West Indies | 11.76% | 6 |
| Zimbabwe | 5.08% | 7 |

Table VI presents the unpredictability index of teams when they won a match while defending a target. This shows how many times a team actually won the match while defending when D/L formula was predicting against it at 40th over.

TABLE VI. D/L FORMULA FAILING PERCENTAGE AT 40TH OVER WHEN A TEAM WON THE MATCH WHILE DEFENDING A TARGET

| Team | Percentage | Rank |
|---|---|---|
| England | 21.95% | 1 |
| India | 17.65% | 2 |
| SriLanka | 13.89% | 3 |
| Pakistan | 11.43% | 4 |
| Australia | 8.16% | 5 |
| South Africa | 7.27% | 6 |
| New Zealand | 7.14% | 7 |

Winning the match unpredictably when defending a target is either because of captainship skills or good death bowling. India is considered to have great captainship skills with number 2 while England tops the list with the mixture of both good captainship and good death bowling.

Table VII presents the unpredictability index of teams when they lost a match while defending a target. This shows how many times a team actually lost the match while defending when the D/L formula was predicting against it.

TABLE VII. D/L FORMULA FAILING PERCENTAGE AT 40TH OVER WHEN A TEAM LOST THE MATCH WHILE DEFENDING A TARGET

| Team | Percentage | Rank |
|---|---|---|
| Australia | 20.41% | 1 |
| England | 17.65% | 2 |
| India | 15.22% | 3 |
| New Zealand | 14.58% | 4 |
| West Indies | 13.64% | 5 |
| SriLanka | 8.33% | 6 |

It has been seen in ODI cricket a number of times where Australian bowlers give plenty of runs at the end to lose the match when defending a target which is evident from Table VII.

## IV. CONCLUSIONS

The paper compared the performance of a few classifiers to predict the outcome of an ODI cricket match during second innings against the D/L formula and found that the machine learning algorithms performed better. It is obvious, however, that the trained classifiers lack the ability to describe the pattern of prediction and are mainly useful for predicting the right winner of a cricket match. A big leap in prediction accuracy by our proposed models is at 10 overs during second innings. In this instance, our classifiers predict the result of the match with slightly more accuracy than what the D/L formula can achieve after 20 overs. Thus, making our proposed classifiers much more effective if the result is to be decided before 20 overs. This high accuracy in predicting the winner during the early phase of the game is always worthwhile from the betting perspective.

DLS Resource Table was also optimized using particle swarm optimization to give better results which shows that there is room for improvement in the table values. We also found that the decreasing nature of the table values when early wickets have fallen should be reanalyzed to devise a better table.

Finally, our Unpredictability Index shows interesting patterns that coincide with the views of a number of experienced cricket analysts.

Match - https://www.icc-cricket.com/about/cricket/rules-and-regulations/duckworth-lewis-stern

[11] Wikipedia: About the Twenty20 form of cricket http://en.wikipedia.org/wiki/Twenty20.